# ATTENTION-BASED MULTIPLE INSTANCE LEARNING FOR PREDOMINANT GROWTH PATTERN PREDICTION IN LUNG ADENOCARCINOMA WSI USING FOUNDATION MODELS

*Laura Valeria Perez-Herrera*[*±]     *M.J. Garcia-Gonzalez*[*±]     *Karen Lopez-Linares*[*±]

\* Vicomtech Foundation, Basque Research and Technology Alliance (BRTA)
± eHealth Group, Bioengineering Area, Biogipuzkoa Health Research Institute

## ABSTRACT

Lung adenocarcinoma (LUAD) grading depends on accurately identifying growth patterns, which are indicators of prognosis and can influence treatment decisions. Common deep learning approaches to determine the predominant pattern rely on patch-level classification or segmentation, requiring extensive annotations. This study proposes an attention-based multiple instance learning (ABMIL) framework to predict the predominant LUAD growth pattern at the whole slide level to reduce annotation burden. Our approach integrates pretrained pathology foundation models as patch encoders, used either frozen or fine-tuned on annotated patches, to extract discriminative features that are aggregated through attention mechanisms. Experiments show that fine-tuned encoders improve performance, with Prov-GigaPath achieving the highest agreement ($\kappa$ = 0.699) under ABMIL. Compared to simple patch-aggregation baselines, ABMIL yields more robust predictions by leveraging slide-level supervision and spatial attention. Future work will extend this framework to estimate the full distribution of growth patterns and validate performance on external cohorts.

## 1. INTRODUCTION

Lung adenocarcinoma (LUAD), the most common subtype of non–small cell lung cancer (NSCLC), presents diverse histologic growth patterns that reflect tumor differentiation and prognosis. According to the World Health Organization classification, LUAD comprises five major patterns: lepidic, acinar, papillary, micropapillary, and solid [1]. The relative proportion of these patterns is clinically relevant, as lepidic-predominant tumors are associated with favorable outcomes, whereas solid and micropapillary patterns indicate poor prognosis. Accordingly, the International Association for the Study of Lung Cancer grading system classifies LUAD by the predominant pattern and the presence of high-grade components [2].

Because pattern assessment is subjective and prone to interobserver variability, artificial intelligence (AI) can improve its reproducibility and objectivity. Previous studies have mainly relied on patch-level classification or segmentation with simple slide-level aggregation. For example, [3] used segmentation-based aggregation to predict tumor grade ($\kappa$ = 0.56), while [4] trained a patch-level classifier with a tailored aggregation strategy to predict the predominant growth pattern at the slide level ($\kappa$ = 0.52). In contrast, [5] reported patch-level performance for distinguishing individual growth patterns (macro-F1 = 0.70) without evaluating slide-level predictions. Although these studies demonstrated the feasibility of patch-based methods, their reliance on densely annotated datasets and simple aggregation strategies (e.g., majority voting) limits their scalability and ability to capture whole-slide spatial context.

Multiple instance learning (MIL) addresses these limitations by using slide-level labels, allowing models to identify discriminative regions without exhaustive annotation. ABMIL further enhances interpretability by assigning attention weights to regions, highlighting those most relevant ones for slide-level prediction [6].

Thus, we employ ABMIL for slide-level classification of the predominant LUAD growth pattern in whole slide images (WSIs). To obtain representative patch-level features, pathology foundation models are used as feature extractors in both frozen and fine-tuned settings. Although these models have demonstrated strong transfer capabilities, their potential for LUAD growth pattern classification remains underexplored. To assess the value of slide-level supervision, ABMIL is compared with a baseline that aggregates patch-level predictions via majority voting, as in prior studies.

## 2. MATERIALS AND METHODS

### 2.1. Datasets

Two publicly available datasets were used in this study with complementary purposes: the ANORAK dataset [3] to fine-tune pathology foundation models at the patch level, and the Dartmouth Lung Cancer Histology (DHMC) dataset [4] to train and evaluate ABMIL models for WSI classification.

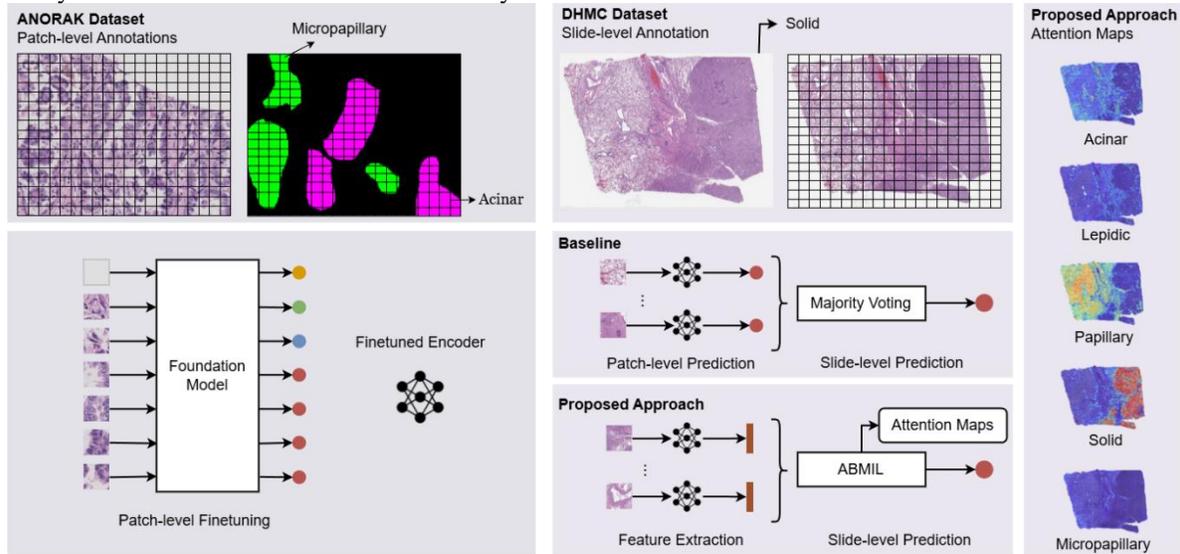

**Fig. 1.** Overview of the proposed framework. Patch-level features extracted from foundation models (frozen or fine-tuned on ANORAK) are used to train an ABMIL model for slide-level prediction on DHMC, while a simpler baseline based on majority voting serves as comparison. The right column shows attention maps produced by the ABMIL model, which highlight the regions that contribute most to each predicted growth pattern.

*2.1.1. ANORAK dataset*

This dataset comprises 731 regions of interest (ROIs) from 49 Hematoxylin and Eosin (H&E) WSIs, derived from the multi-center TRACERx100 cohort. Each ROI was independently annotated at ≈0.45μm/pixel (20× magnification) by three pathologists, providing pixel-level labels for six LUAD growth patterns: lepidic, acinar, papillary, solid, micropapillary, and cribriform, as well as non-tumor regions. From these ROIs, non-overlapping 448×448 patches were extracted with patch-level labels derived from the pixel annotations by retaining only those entirely covered by a single growth pattern. Cribriform patches were excluded, as this subtype is not among the five major patterns defined by the current WHO classification. Only full-size patches within annotated ROIs were retained, and patches containing less than 10% tissue were discarded.

This approach yielded 4,085 patches used to fine-tune pathology foundation models in a fully supervised patch-level setting. To prevent data leakage, dataset splitting was performed at patient level, resulting in 3,226 training and 859 validation patches, distributed as follows: background (1262/359), lepidic (240/121), acinar (663/146), papillary (303/51), micropapillary (38/10) and solid (720/172).

*2.1.2. DHMC dataset*

The DHMC dataset has 143 H&E WSIs from the Department of Pathology and Laboratory Medicine at Dartmouth–Hitchcock Medical Center, each labeled with the predominant growth pattern by the consensus of three pathologists. Class counts are: lepidic (19), acinar (59), papillary (5), solid (51), and micropapillary (9). Images were analyzed at 20× magnification, and tissue patches were extracted following the CLAM methodology [6]. This dataset was used to train and evaluate the ABMIL models for slide-level classification, using features extracted from frozen or fine-tuned encoders. Given the limited sample size, a five-fold cross-validation scheme was adopted to ensure robust performance estimation.

### 2.2. Methodology

Figure 1 shows the overall workflow, which integrates patch-level fine-tuning of pathology foundation models to make the slide-level prediction through an ABMIL framework.

*2.2.1. Foundation models and patch-level fine-tuning*

Five pathology foundation models were evaluated as frozen feature extractors and after patch-level fine-tuning with the ANORAK dataset. Additionally, a standard ViT-L/16 [7] pretrained on ImageNet [8] was included as a non–pathology-specific baseline.

Key architectural and training characteristics of the evaluated models are summarized in Table 1, and readers are referred to the original publications for comprehensive implementation details. For models with multimodal or hierarchical components, such as CONCH (which integrates a text encoder) and Prov-GigaPath (which includes an additional slide-level encoder), only the specifications relevant to the patch encoder are reported here.

Each vision encoder was fine-tuned for patch-level classification using images normalized with the ImageNet mean and standard deviation. Data applied using *albumentations* [13], including horizontal and vertical flips, 90° rotations, hue–saturation–value shifts, random brightness and contrast adjustments, Gaussian blur, and JPEG compression. Training employed weighted sampling to mitigate class imbalance, the Adam optimizer with a learning rate of $1\times10^{-5}$, batch size 8, and early stopping with a patience of 15 epochs.

**Table 2.** Summary of evaluated foundation models. EMB = embedding dimension.

| FM | Architecture | Params | Patches | Slides | Pretraining | EMB | Lung Proportion |
|---|---|---|---|---|---|---|---|
| UNIv1 [9] | ViT-L/16 | 303M | 100M | 100K | DINOv2 | 1024 | ≈9% |
| UNIv2 [9] | ViT-H/14 | 681M | 200M | 350K | DINOv2 | 1536 | Nor specified |
| Virchow2 [10] | ViT-H/14 | 631M | 2B | 3.1M | mDINOv2 | 2560 | ≈4% |
| CONCH [11] | ViT-B/16 | 90M | 1.17M | - | CoCa | 768 | Third |
| Prov-GigaPath [12] | ViT-G/14 | 1.1B | 1.3B | 171K | DINOv2 | 1536 | 45.29% |

**Table 2.** Results on the **DHMC** dataset for foundation models evaluated under: (1) Patch-level classification using **majority voting** for aggregation, and (2) **ABMIL classification**. **Bold** and underline indicate best and second best results, respectively.

| | | (1) Majority voting | | | (2) ABMIL | |
|---|---|---|---|---|---|---|
| | FM | Weighted F1 | $\kappa$ | FM | Weighted F1 | $\kappa$ |
| Frozen | ViT-L | 0.338±0.010 | 0.260±0.017 | ViT-L | 0.594±0.075 | 0.419±0.108 |
| | UNIv1 | 0.541±0.052 | 0.397±0.050 | UNIv1 | 0.638±0.063 | 0.505±0.106 |
| | UNIv2 | 0.567±0.047 | 0.401±0.065 | UNIv2 | 0.693±0.051 | 0.553±0.084 |
| | Virchow2 | 0.577±0.052 | 0.411±0.075 | Virchow2 | 0.697±0.100 | 0.557±0.151 |
| | CONCH | 0.486±0.061 | 0.324±0.057 | CONCH | 0.704±0.077 | 0.578±0.112 |
| | Prov-GigaPath | 0.547±0.057 | 0.392±0.061 | Prov-GigaPath | 0.671±0.080 | 0.524±0.109 |
| Fine-tuned | ViT-L | 0.651±0.058 | 0.492±0.096 | ViT-L | 0.684±0.132 | 0.562±0.169 |
| | UNIv1 | 0.655±0.041 | <u>0.521±0.062</u> | UNIv1 | <u>0.725±0.065</u> | <u>0.606±0.093</u> |
| | UNIv2 | <u>0.656±0.021</u> | **0.539±0.023** | UNIv2 | 0.672±0.113 | 0.542±0.145 |
| | Virchow2 | 0.639±0.034 | 0.483±0.031 | Virchow2 | 0.433±0.156 | 0.243±0.201 |
| | CONCH | 0.648±0.069 | 0.487±0.089 | CONCH | <u>0.726±0.072</u> | 0.597±0.114 |
| | Prov-GigaPath | **0.663±0.053** | 0.508±0.099 | Prov-GigaPath | **0.788±0.062** | **0.699±0.086** |

*2.2.2. ABMIL*

The patch encoders described above served as feature extractors within the ABMIL framework for slide-level classification. For each WSI in the DHMC dataset, tissue patches were first embedded into high dimensional feature vectors using the selected encoder (frozen or fine-tuned on ANORAK). These patch embeddings were then aggregated by the ABMIL model to predict the predominant growth pattern of the slide.

We implemented a CLAM style ABMIL architecture with class-specific gated attention and per class bag classifiers. Each patch feature vector was projected into a 512-dimensional embedding through a linear layer. For each class, a gated-attention mechanism computed normalized attention scores across tiles, which are aggregated into class-specific bag representations. Independent linear heads then mapped these to scalar logits, which are concatenated to produce the final multi-class prediction. The model was trained using five-fold cross-validation, employing a cross-entropy loss, the Adam optimizer (learning rate $1\times10^{-4}$), batch size of 1, and weighted sampling. Early stopping was applied based on validation loss.

## 3. RESULTS AND DISCUSSION

### 3.1. Patch-level fine-tuning

At the patch level, frozen encoders achieved good agreement with expert annotations ($\kappa > 0.965$) across all foundation models, indicating that pretrained representations already capture highly discriminative local features. Fine-tuning on ANORAK further improved performance, reaching near perfect agreement ($\kappa > 0.970$) and confirming the models' strong capacity to distinguish LUAD morphologies. Among all models, Prov-GigaPath achieved the highest validation ($\kappa$ = 0.993), closely followed by UNIv1 and Virchow2, while the ImageNet baseline (ViT-L) performed worst (0.970). While the training curves did not indicate overfitting, these high scores are partly due to the task being simplified: each patch contained a single growth pattern within its field of view. In contrast, other studies typically evaluate more heterogeneous patches or perform pixel-level segmentation (e.g., [3]). This simplification was intentional, as our objective was to adapt the encoders for later slide-level prediction rather than optimal patch-level performance.

## 3.2. Slide-level performance

After fine-tuning the models at patch level, they were used to extract features from the DHMC dataset for training ABMIL models to aggregate patch information and predict the predominant growth pattern at the slide level. The results for both majority voting aggregation and ABMIL strategies, using frozen and ANORAK fine-tuned encoders, are summarized in Table 2.

When aggregating patch-level predictions into slide-level classifications using simple majority voting, frozen encoders showed variable performance across architectures (as shown in Table 2 (1)). As expected, ViT-L (ImageNet baseline) performed the worst, confirming the advantage of pathology specific pretraining. Among pathology foundation models, Virchow2 and UNIv2 achieved the best results (weighted F1 $\geq$ 0.567, $\kappa \geq$ 0.401), followed by Prov-GigaPath and UNIv1, while CONCH showed the lowest performance. These trends likely reflect differences in model scale and pretraining data where Prov-GigaPath, Virchow2 and UNIv2 are the largest models trained on a greater number of image patches. Prov-GigaPath's slightly lower performance despite its size may be explained by the fact that its dedicated whole slide encoder was not used during inference, potentially limiting the model's ability to fully leverage slide-level contextual information.

After fine-tuning on ANORAK, all models, including ViT-L, converged to similar slide-level performance, reflecting the convergence already seen during patch-level training. Prov-GigaPath achieved the highest weighted F1 (0.663) and UNIv2 the highest $\kappa$ score (0.539), closely followed by UNIv1. Notably, ViT-L reached performance comparable to or exceeding some foundation models (Virchow2 and CONCH), suggesting that patch-level fine-tuning on the target task can compensate for the lack of pathology pretraining. However, even if fine-tuning improved performance, majority voting aggregation ignores the spatial organization of growth patterns within slides. To better capture LUAD heterogeneity, we next evaluated ABMIL, which incorporates slide-level supervision to weight patch contributions adaptively.

Under the ABMIL setting, frozen encoders showed a different behavior (Table 2 (2)), with CONCH achieving the highest weighted F1 (0.704), followed by Virchow2 and UNIv2. This suggests that CONCH's vision–language pretraining, which aligns image and text representations, can yield transferable histopathology features even with smaller model capacity, particularly when coupled with a slide-level mechanism such as ABMIL. In contrast, Prov-GigaPath, despite its large scale and abundant lung tissue representation in pretraining, did not achieve top results in the frozen setting, confirming its reliance on the original slide encoder to integrate contextual information, consistent with the majority voting results.

After fine-tuning, nearly all foundation models improved, as expected. Prov-GigaPath achieved the best overall results (F1 = 0.788, $\kappa$ = 0.699), significantly outperforming all models (p $\leq$ 0.012) but UNIv1 (p = 0.079), one of the next best performing models. These gains confirm that its representations are highly adaptable and benefit from task specific optimization. Conversely, Virchow2's marked decline may be due to hyperparameter sensitivity or from the compression imposed by the ABMIL projection layer, reducing its feature dimensionality (from 2,560 to 512) and potentially the richness of its learned representations.

To gain deeper insight into model behavior, per-class F1-scores were analyzed. Performance was substantially lower for minority classes, with the most pronounced degradation observed for the severely underrepresented papillary subtype (F1-score $\leq$ 0.467) across all models and aggregation strategies. Under the fine-tuned ABMIL setting, among the top-performing models, Prov-GigaPath exhibited the most balanced behavior, achieving non-zero F1-scores for all histological subtypes. In contrast, CONCH failed to predict the papillary subtype, while UNIv1 showed reduced performance for the lepidic class (F1-score 0.526 vs. 0.733 for Prov-GigaPath).

Overall, performance trends varied across foundation models, but task-specific fine-tuning generally improved results across both strategies (p $\leq$ 0.0442), with ABMIL yielding the best performance. While a direct comparison with patch-level aggregation is not strictly equivalent (given that ABMIL incorporates domain-specific priors from the target dataset) the results demonstrate that, when coupled with limited training data, slide-level supervision can achieve competitive and clinically meaningful performance, with reduced annotation burden. These results highlight the potential of weakly supervised methods in computational pathology, even without exhaustive patch-level annotations.

## 4. CONCLUSIONS AND FUTURE DIRECTIONS

This study presents an ABMIL framework for predicting the predominant growth pattern in LUAD WSIs using pathology foundation models as patch-level encoders. While both ABMIL and conventional patch aggregation benefited from pretrained encoders, task-specific fine-tuning further improved performance. Overall, ABMIL proved more effective than simple patch aggregation approaches commonly used in prior studies.

Future work will extend this analysis by leveraging ABMIL attention mechanisms to estimate the spatial distribution of growth patterns within each slide, beyond the predominant type; conducting external validation on independent LUAD cohorts to assess cross-institutional generalizability; and incorporating the full Prov-GigaPath architecture, including the slide-level encoder. In addition, model-specific hyperparameter optimization and parameter-efficient fine-tuning strategies will be explored to improve cross-cohort robustness while preserving computational efficiency.

## 5. COMPLIANCE WITH ETHICAL STANDARDS

This research study was conducted retrospectively using human subject data generated and made available in open access by ANORAK dataset (*https://zenodo.org/records/10016027*). Ethical approval was not required as confirmed by the license attached with the open access data. DHMC dataset (*https://bmirds.github.io/LungCancer/*) was de-identified and released with permission from Dartmouth-Hitchcock Health (D-HH) Institutional Review Board.

## 6. ACKNOWLEDGEMENTS

Funded by the European Union's Horizon Europe Research and Innovation Programme under Grant Agreement no. 101096473. Views and opinions expressed are, however, those of the author(s) only and do not necessarily reflect those of the European Health and Digital Executive Agency. Neither the European Union nor the granting authority can be held responsible for them.